\pdfoutput=1

\documentclass[11pt]{article}
\usepackage[]{authblk}
\usepackage[]{acl}

\usepackage{times}
\usepackage{latexsym}

\usepackage[T1]{fontenc}

\usepackage[utf8]{inputenc}

\usepackage{microtype}
\usepackage{amsfonts}
\usepackage{amsthm}
\theoremstyle{definition}
\newtheorem{definition}{Definition}[section]
\usepackage{csquotes}
\usepackage{amsmath}
\usepackage{multirow}
\usepackage{graphicx}
\usepackage{enumitem}
\usepackage{booktabs}
\usepackage{relsize}

\newcommand*{\Girish}[1]{\textcolor{purple}{[Girish: #1]}}

\title{Synthetic Text Generation with Differential Privacy: \\A Simple and Practical Recipe}

\author[1,\thanks{\ \  Most of the work was done when Xiang, Xuechen, and Girish interned at Microsoft (Research).}]{Xiang Yue}
\author[2]{Huseyin A. Inan}
\author[3]{Xuechen Li}
\author[5]{\\Girish Kumar}
\author[4]{Julia McAnallen}
\author[4]{Hoda Shajari}
\author[1]{Huan Sun}
\author[4]{David Levitan}
\author[2]{Robert Sim}

\makeatletter
\renewcommand\AB@affilsepx{, \protect\Affilfont}
\makeatother
\affil[1]{The Ohio State University}
\affil[2]{Microsoft Research}
\affil[3]{Stanford University}
\affil[4]{Microsoft}
\makeatletter
\renewcommand\AB@affilsepx{\\ \protect\Affilfont}
\makeatother
\affil[5]{UC Davis}

\affil[ ]{\relscale{0.87}\{\texttt{yue.149,sun.397\}@osu.edu}}
\affil[ ]{\relscale{0.87}\texttt{lxuechen@cs.stanford.edu} \quad \texttt{gkum@ucdavis.edu}}
\affil[ ]{\relscale{0.87}\{\texttt{Huseyin.Inan,Julia.McAnallen,hodashajari,David.Levitan,rsim\}@microsoft.com}}
\begin{document}
\maketitle

\begin{abstract}
Privacy concerns have attracted increasing attention in data-driven products due to the tendency of machine learning models to memorize sensitive training data. Generating synthetic versions of such data with a formal privacy guarantee, such as differential privacy (DP), provides a promising path to mitigating these privacy concerns, but previous approaches in this direction have typically failed to produce synthetic data of high quality. In this work, we show that a simple and practical recipe in the text domain is effective: simply fine-tuning a pre-trained generative language model with DP enables the model to generate useful synthetic text with strong privacy protection. Through extensive empirical analyses on both benchmark and private customer data, we demonstrate that our method produces synthetic text that is competitive in terms of utility with its non-private counterpart, meanwhile providing strong protection against potential privacy leakages.\footnote{Our code is available at \url{https://github.com/microsoft/dp-transformers}}
\end{abstract}

\section{Introduction}
The issue of privacy has gained increasing attention in natural language processing (NLP). Privacy attacks against common NLP pipelines have demonstrated that models trained without formal privacy guarantees can reveal membership information and enable training data reconstruction~\cite{shokri2017membership,carlini2021extracting}.
Privacy concerns manifested through tightening legislation~(e.g., GDPR \cite{GDPR}) and growing discussions on policy and ethics call for improved approaches for privacy-preserving machine learning.

\begin{figure*}[!t]
    \centering
    \includegraphics[width=0.99\linewidth]{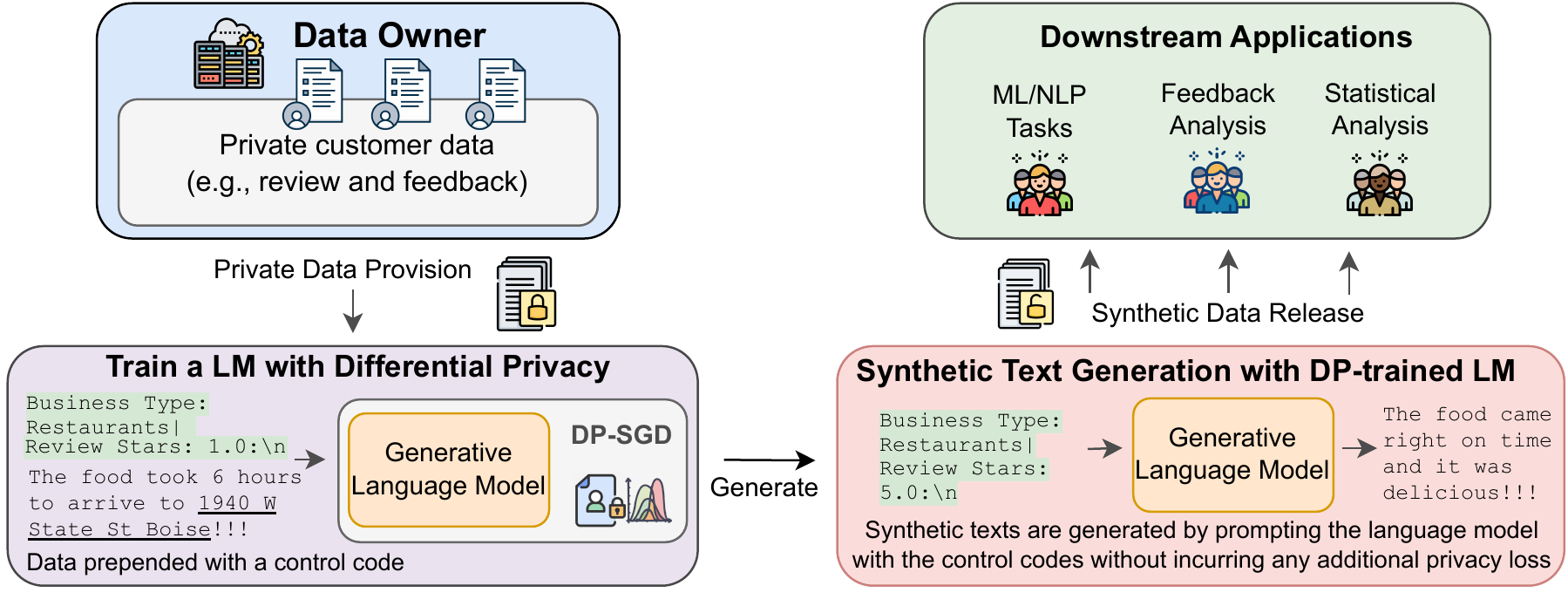}
    \caption{Illustration of our problem and methodology. 
    We propose to generate synthetic text with a formal privacy guarantee: we fine-tune a generative language model with DP and then leverage it for synthetic text generation using control codes. Privacy loss of the overall procedure can be controlled by the data generation stage as, by the robustness to post-processing property of DP, the downstream task stage does not incur any additional privacy loss.}
    \label{fig:method}
\end{figure*}

Among different approaches for learning with private data, learning with differential privacy (DP)~\cite{dwork2006calibrating} has become the gold standard as its formal guarantee enables reasoning about the privacy loss in a principled manner and makes the approach resilient to strong privacy attacks~\cite{carlini2019secret}. 
Recent developments have substantially improved the computational efficiency and privacy-utility trade-off of DP machine learning~\cite[][\emph{inter alia}]{subramani2021enabling,li2022large,yu2022differentially,de2022unlocking,bu2022scalable,li2022does,mehta2022large}, demonstrating gains for learning models that perform specific downstream tasks. 

In contrast to the above works, we study \textit{synthetic text generation by building generative text models with DP training algorithms} (Figure \ref{fig:method}).
The goal of this approach is to learn a generative model that faithfully captures distributional properties of the training data (and the underlying distribution), as opposed to learning task-oriented models with specific functions. 
Compared to directly learning models for target tasks, this paradigm has several advantages: 
(1) DP-trained generative models can be used to draw synthetic data for learning an expanding set of task models without incurring any additional privacy loss (due to the post-processing property of DP);
(2) Dataset debugging is made easy as synthetic text generated from DP-trained models can be shared more freely, and inspecting its samples poses less of a privacy concern compared to examining the original private data~\cite{augenstein2019generative}; 
(3) Synthetic data generated from DP-trained models can be retained for a longer time under certain existing policies (e.g., \textit{right to be forgotten}) thanks to the fact that DP implies some degree of approximate machine unlearning~\cite{bourtoule2021machine,sekhari2021remember}. 

In this work, we initiate a systematic empirical study of the problem and show that DP language model (LM) fine-tuning can be an effective solution to synthetic text generation with privacy.
In particular, we show that simply fine-tuning progressively larger autoregressively pre-trained language models on (private) data leads to models that generate increasingly useful synthetic text. 
For instance, we fine-tune a GPT-2 Large model \cite{radford2019language} on a review dataset with DP at $\epsilon=4$ and then use it to generate synthetic text to build downstream classifiers. 
The classification models achieve comparable performance (only 2-4\% in accuracy away) to the classifiers trained on the original dataset.

Furthermore, we demonstrate that generating a small amount of synthetic data with DP is sufficient to create classification models that are on par with those trained directly on the entire original dataset with DP. One of the advantages of the synthetic data approach is that the privacy loss is fixed, and an unlimited number of downstream models can be built without incurring additional leakage. In contrast, training additional downstream models on the original data with DP accumulates privacy loss.

Distributional similarity evaluation additionally confirms that the synthetic text distribution resembles the original data distribution. We also uncover a novel phenomenon in DP-trained LMs that is of independent interest. Specifically, we observe a \emph{length truncation effect} in text generation with DP-trained models, resulting in completions that are generally shorter than their non-DP counterparts and instances in the original dataset.

We further extensively study learning dynamics with DP by injecting specially-crafted \emph{canaries}~\cite{carlini2019secret} in the training data. This allows for (i) stress-testing the extent to which DP fine-tuning limits the \textit{leakage of private information} and (ii) understanding the conditions under which a \textit{subject of interest} would appear in synthetic generations.

Finally, we conclude our studies on an industrial-level private customer feedback dataset to show the feasibility of our approach in real-world scenarios. 

\section{Background}
\subsection{Differential Privacy}
\label{sec:DP_def}
\begin{definition}[Differential Privacy (DP)~\cite{dwork2006calibrating}]
\label{def:dp}
A randomized algorithm $M: \mathcal{D} \rightarrow \mathcal{S}$ is  $(\epsilon,\delta)$-differentially private if for any two neighboring datasets $D, D' \in \mathcal{D}$ that differ exactly in a single data sample, and for all sets $S\subseteq\mathcal{S}$:
\begin{equation*}
        \mathbb{P}[M(D) \in S] \leq e^{\epsilon} \mathbb{P}[M(D') \in S] + \delta.
\end{equation*}
\end{definition}
This definition provides a rigorous privacy guarantee by theoretically bounding the effect of a single data sample in the dataset. For a differentially private algorithm, the output distribution is statistically similar whether any individual data sample appears in the input dataset or not. The privacy parameter $\epsilon$ quantifies the maximum allowable impact of a single individual's data on the outcome. $\delta$ specifies the maximum probability that the privacy guarantee may fail. An algorithm can typically be made $(\epsilon, \delta)$-DP by bounding the contribution of a single data sample and adding controlled noise from a predetermined distribution (e.g., Gaussian)~\cite{dwork2014algorithmic}. Setting $\epsilon$ and $\delta$ in practice often requires careful consideration of the specific use case and the acceptable trade-off between privacy and utility. We discuss our choice of $\epsilon$ and $\delta$ in Section \ref{sec:exp_setup}.

An appealing property of DP crucial to this work is \textit{robustness to post-processing}. This property ensures that if the algorithm $M$ satisfies $(\epsilon, \delta)$-DP, then so does $F \circ M$ for any deterministic or randomized function $F$ (which is independent of $M$). Namely, one can perform arbitrary post-processing without incurring additional privacy loss.

\subsection{DP Stochastic Gradient Descent}
Deep learning models can be trained with DP via a modification of the stochastic gradient descent (SGD) algorithm~\cite{song2013stochastic, BassilyST14, ccs/AbadiCGMMT016}. 
The modified algorithm clips \textit{per-sample gradients} to bound the contribution of individual examples. 
Noise from a Gaussian distribution is sampled and added to the sum of the clipped gradients in a batch to obfuscate the gradient update. 
The resulting algorithm, called Differentially Private Stochastic Gradient Descent (DP-SGD), can be shown to be DP for some $(\epsilon, \delta)$ for each update of the model. Privacy parameters at the end of training can be computed via privacy composition algorithms \cite{ccs/AbadiCGMMT016,nips/GopiLW21}. 
In the next section, we will utilize DP-SGD to train a language model with privacy for synthetic text generation.

\section{Method}
In this section, we formally state the problem and present our method (see Figure \ref{fig:method} for an illustration) that produces a synthetic version of private text data with differential privacy.

\subsection{Problem Statement}
\label{sec:prob_state}
Let $\mathcal{D}$ be a database representing the collection of token sequences from a fixed dictionary $\mathcal{V}$. We define a (randomized) mapping $M: \mathcal{D} \rightarrow \mathcal{D}$ such that for a given dataset $D \in \mathcal{D}$, the goal is to generate a synthetic version $M(D) = \Tilde{D}$ with privacy constraints and utility desiderata.

Regarding privacy constraints, we require that $M$ be $(\epsilon, \delta)$-DP with domain $\mathcal{D}$. This requirement provides strong protection for the participants in the input dataset as this participation will be statistically indistinguishable to a certain degree through any adversary accessing the model or synthetic version of the dataset in the output.

For the case of utility, ideally, the synthetic version $\Tilde{D}$ should be able to replace $D$ in providing a training resource for models on relevant downstream applications. In other words, on target downstream tasks, models trained on the synthetic dataset $\Tilde{D}$ are expected to have performance similar to the models trained on the original dataset $D$. More generally, distributional properties of the dataset $D$ should be captured as much as possible in the synthetic version $\Tilde{D}$ without violating the aforementioned privacy requirement. These will be extensively explored in Section \ref{sec:main_exp}.

\subsection{Synthetic Text Generation with DP}

Conventionally, to generate synthetic text, an auto-regressive language model (e.g. GPT-2~\cite{radford2019language}) is trained on the original dataset and subsequently sampled using a sampling mechanism (e.g., beam search, top-$k$ sampling~\cite{fan-etal-2018-hierarchical}, nucleus sampling~\cite{iclr/HoltzmanBDFC20}, etc.) to produce synthetic sequences. 

To make this operation differentially private, we adopt DP-SGD to fine-tune a pre-trained generative LM. The post-processing property of DP ensures that once the LM has been fine-tuned with DP, sampling from the model incurs no extra privacy loss.


It would be desirable to synthesize examples with labels. We achieve this by building a conditional generator introduced in \cite{Keskar19} to provide more explicit control over text generation. By using so-called control codes \cite{Keskar19}, the probability distribution of a text sequence $x = \left(x_1, x_2, \ldots, x_n\right)$ is conditioned on a control code $c$ and decomposed as: 
\begin{equation*}
    \mathbb{P} \left(x|c\right) = \prod_{i=1}^{n} \mathbb{P} \left(x_i | x_1, x_2, \ldots, x_{i-1}, c\right).
\end{equation*}
\vspace{-5pt}

A neural network $p_\theta(\cdot)$ is then trained to model each conditional distribution. The model can later be used to generate new samples conditioned on a control code $c$ by sequentially sampling $p_\theta(x_1 | c), p_\theta(x_2 | \tilde{x}_1, c), \ldots, p_\theta(x_m | \tilde{x}_1, \ldots \tilde{x}_{m-1}, c)$. The advantage of this approach is that it provides flexibility in the text generation of the model by allowing the conditional control codes to specify a particular style, domain, sentiment, or category. 
For example, feedback data collected from users on a set of products may contain product types and review scores associated with each data sample. Control codes can be constructed as $c_{p, r} =$ \textit{"Product type: $p$ | Review score: $r$"} for different product type ($p$) and review score ($r$) pairs. In our method, we utilize control codes to prepend each sample with its corresponding categories as a simple preprocessing step. During the text generation, this allows us to use the control codes to generate as many samples as the original categorical distribution is preserved.

We point out that the categorical distribution in the original dataset may also be a piece of private information itself. However, its estimation could easily be privatized~\cite{dwork2014algorithmic} and for simplicity, we ignore the low-cost privacy loss of this step and use the exact categorical distribution of the original dataset in this paper.
\section{Analyses on a Public Review Dataset}
\label{sec:main_exp}
In this section, we extensively analyze our method with experiments on a public benchmark dataset: Yelp Open Dataset,\footnote{\url{https://www.yelp.com/dataset}} which has been widely adopted for language modeling and text classification tasks. 
We then apply our method to an internal private customer feedback dataset in Section \ref{sec:case_study_feedback}.

\subsection{Experimental Setup}
\label{sec:exp_setup}
\paragraph{Dataset.} The Yelp dataset contains review text data on businesses that can be studied for academic purposes. We select two attributes for the conditional generation as well as the downstream task applications: review stars (1-5) and business category. We sample 10 frequent business categories and remove the reviews that do not have ratings (Details can be found in Appendix \ref{apdx:sec:yelp_dataset}).
This results in a dataset that has 1.9M reviews for training, 5000 for validation, and 5000 for testing. 

\paragraph{Implementation Details.} We utilize the public repository~\cite{dp-transformers}, which is based on Huggingface~\cite{huggingface} and Opacus~\cite{opacus}, for fine-tuning language models with DP. Specifically, we fine-tune three language models: GPT2~\cite{radford2019language}, GPT2-Medium, and GPT2-Large, for synthetic text generation. Additionally, we fine-tune the RoBERTa-base model~\cite{corr/abs-1907-11692} for downstream text classification tasks.

Control codes are constructed based on attributes such as \textit{``Business Type: Bar | Review Stars: 5.0''} and are prepended to each sample. Hyperparameters are specified in Appendix \ref{sec:hyperparams}. For both synthetic text generation and classification, we set the maximum sequence length to 128, unless otherwise specified. During training, we evaluate the models on the dev dataset and select the checkpoint that achieves the best validation performance for the final evaluation on the test set.

We set the privacy parameter $\epsilon$ to 4, which is supported by prior work \cite{yu2021not, li2022large, yu2022differentially, de2022unlocking, mehta2022large} and real-world applications. For instance, the release of US population data uses $\epsilon=13.64$ \cite{Uscensusbureau}, and the development of a next-word prediction model uses $\epsilon=6.92$ \cite{googlefl}. Our $\epsilon=4$ is smaller and provides stronger privacy protection.  As recommended by \cite{hsu2014differential, de2022unlocking}, $\delta$ should be smaller than the inverse of the dataset size $N$, and we set $\delta=1/(N \cdot \log N)$. The additive noise scale is calculated using the numerical composition algorithm \cite{gopi2021numerical}, given the batch size and epochs for each setting mentioned in Appendix \ref{sec:hyperparams} for DP training.

\begin{table}[!t]
\centering
\resizebox{\linewidth}{!}{%
\begin{tabular}{llccc}
\toprule
Data Type & Data Generator & $\epsilon$ & Rating & Category \\ \midrule
Original  & - & - & 0.7334 & 0.7752 \\ \midrule
\multirow{6}{*}{Synthetic} & \multirow{2}{*}{GPT2} & $\infty$ & 0.6892 & 0.7584 \\  
 &  & 4 & 0.6656 & 0.7478 \\ \cmidrule(lr){2-5} 
 & \multirow{2}{*}{GPT2-Medium} & $\infty$ & 0.6878 & 0.7550 \\ 
 &  & 4 & 0.6756 & 0.7486 \\ \cmidrule(lr){2-5} 
 & \multirow{2}{*}{GPT2-Large} & $\infty$ & 0.7090 & 0.7576 \\ 
 &  & 4 & 0.6936 & 0.7568 \\ \bottomrule
\end{tabular}
}
\caption{Synthetic text generation with DP yields models that exhibit comparable accuracy in downstream tasks (review rating and business category classification) when compared to models trained on the synthetic text generated without privacy protection.}
\label{tbl:downstream_task_acc}
\vspace{-10pt}
\end{table}

To generate synthetic text samples, we employ top-$k$ sampling \cite{fan-etal-2018-hierarchical} and nucleus sampling (top-$p$) \cite{iclr/HoltzmanBDFC20}, with $k=50$ and $p=0.9$. To produce synthetic datasets that preserve categorical distributions (e.g., business category), we generate 100K samples from the fine-tuned models using the appropriate control codes.

\subsection{Downstream Tasks on Synthetic Data}
\label{sec:downstream_task_acc}
One way to evaluate the quality of the synthetic dataset is by examining the performance of downstream task models trained on it. We fine-tune RoBERTa-base models for classifying review ratings and business categories using the synthetic dataset. We further compare their performance with models trained on the original dataset. All models are evaluated on the same original test set.

The results are summarized in Table \ref{tbl:downstream_task_acc}. The downstream task models trained on the synthetic data generated by GPT2 with DP ($\epsilon=4$) achieve comparable performance to the models trained on the synthetic data generated without DP ($\epsilon=\infty$) and the models trained on the original dataset. Additionally, we observe that the quality of the synthetic generations improves when larger pre-trained language models are used (sampled generations can be found in Appendix \ref{apdx:sampled_generations}), and the performance gap between private and non-private generations diminishes. Surprisingly, models trained on synthetic data generated by GPT2-Large with DP exhibit similar or even better performance compared to models trained on synthetic data generated by GPT2 without DP. These results highlight the significant potential of our method for generating synthetic data across various downstream applications.

\begin{table}[!t]
\centering
\resizebox{\linewidth}{!}{%
\begin{tabular}{ccccc}
\toprule
\multirow{2}{*}{\begin{tabular}[c]{@{}c@{}}Data \\ Type\end{tabular}} & \multirow{2}{*}{\begin{tabular}[c]{@{}c@{}}Data\\ Size\end{tabular}} & \multirow{2}{*}{\begin{tabular}[c]{@{}c@{}}DP \\ Position\end{tabular}} & \multicolumn{2}{c}{Task  Accuracy} \\ \cmidrule(lr){4-5} 
 &  &  & \multicolumn{1}{c}{Rating} & Category \\ \midrule
Original & 1.9M & Task modeling & \multicolumn{1}{c}{0.7014} & 0.7644 \\ \midrule
Original & 100K & Task modeling & \multicolumn{1}{c}{0.6689} & 0.7552 \\ \midrule
Synthetic & 100K & Data Generator & \multicolumn{1}{c}{0.6936} & 0.7568 \\ \bottomrule
\end{tabular}
}
\caption{The model trained on synthetic data generated with DP-trained GPT2-Large (the last row) has similar performance compared to the models directly trained on the original dataset with DP (the first two rows).}
\vspace{-10pt}
\label{tbl:downstream_with_dp}
\end{table}

\subsection{Synthetic Data Generation with DP v.s. Downstream Task Modeling with DP}
It is natural to compare how downstream task models built on synthetic text generated by a DP-trained LM fare against models directly trained on the original data with DP. The results of this comparison are presented in Table \ref{tbl:downstream_with_dp}.

We observe that by using the same privacy parameter ($\epsilon=4$), both approaches achieve comparable performances. However, it is important to note that training two task models on the private dataset with DP will result in a higher overall privacy loss than $\epsilon=4$, and this loss will accumulate with additional downstream tasks. In contrast, the post-processing property of DP allows us to train any number of models for different downstream tasks on the synthetic data generated by a DP-trained LM without incurring additional privacy loss.

An interesting observation is that once the synthetic data is generated with DP, a smaller dataset size (100K instead of 1.9M) is sufficient to produce superior downstream models compared to models directly trained with DP on the original data of the same size (as seen in the second row of Table \ref{tbl:downstream_with_dp}).

\subsection{Similarity between Synth. and Real Data}
\label{sec:similarity_synth_orig}
To further assess the quality of the synthetic generations, we evaluate the similarity between the synthetic dataset and the original dataset. Unlike typical natural language generation tasks like machine translation or summarization, where gold references can be used for evaluation, it is challenging to directly compare synthetic generations with the original dataset when there is no one-to-one mapping between them. In our evaluation, we measure the ``similarity'' from three different perspectives: Embedding Distribution Distance, Topic Difference, and Text Length Distribution.

\begin{table}[!t]
\centering
\small
\resizebox{\linewidth}{!}{%
\begin{tabular}{lcccc}
\toprule
Generator & $\epsilon$ & F1$\uparrow$ & FID$\downarrow$ & MAUVE$\uparrow$ \\ \midrule
\multirow{2}{*}{GPT2} & $\infty$ &0.5199  &3.2368  &0.7158  \\
 & 4 &0.4786  &4.7998  &0.5579  \\ \midrule
\multirow{2}{*}{GPT2-Medium} & $\infty$ &0.5446	 & 3.1464	&  0.7222 \\
 & 4 &0.5076	  &4.1880	 &0.6085   \\ \midrule
\multirow{2}{*}{GPT2-Large} & $\infty$ &0.5852	& 3.0978 & 0.7238\\
 & 4 & 0.5140		 & 4.1352	 & 0.6093	 \\ \bottomrule
\end{tabular}
}
\caption{Distribution distance between the synthetic and original data based on various metrics. Performance improves as larger models are used.}
\vspace{-10pt}
\label{tbl:distribution_distance}
\end{table}

\paragraph{Embedding Distribution Distance.} To measure the embedding distribution distance between the synthetic and original data, we use sentence-transformers \cite{reimers-2019-sentence-bert} to embed both datasets. We calculate the distance between the two distributions using three metrics: 1) F1 Score: the harmonic mean of Precision and Recall \cite{nips/KynkaanniemiKLL19}. Precision estimates the average sample quality, while Recall measures the coverage of the sample distribution. 2) Fréchet Inception Distance (FID): FID calculates the feature-wise mean and covariance matrices of the embedding vectors and then measures the Fréchet distance between the two sets \cite{nips/HeuselRUNH17}. 3) MAUVE: MAUVE compares the distributions of the synthetic and original data using divergence frontiers \cite{nips/PillutlaSZTWCH21}. 

We note that the absolute scale of these metrics may vary depending on the specific embedding models used. To account for this, we conduct the evaluations with five different pre-trained sentence transformers (details provided in Appendix \ref{apdx:embedding_models}), and then compute the average for each metric.

\begin{figure}[!t]
    \centering
    \includegraphics[width=\linewidth]{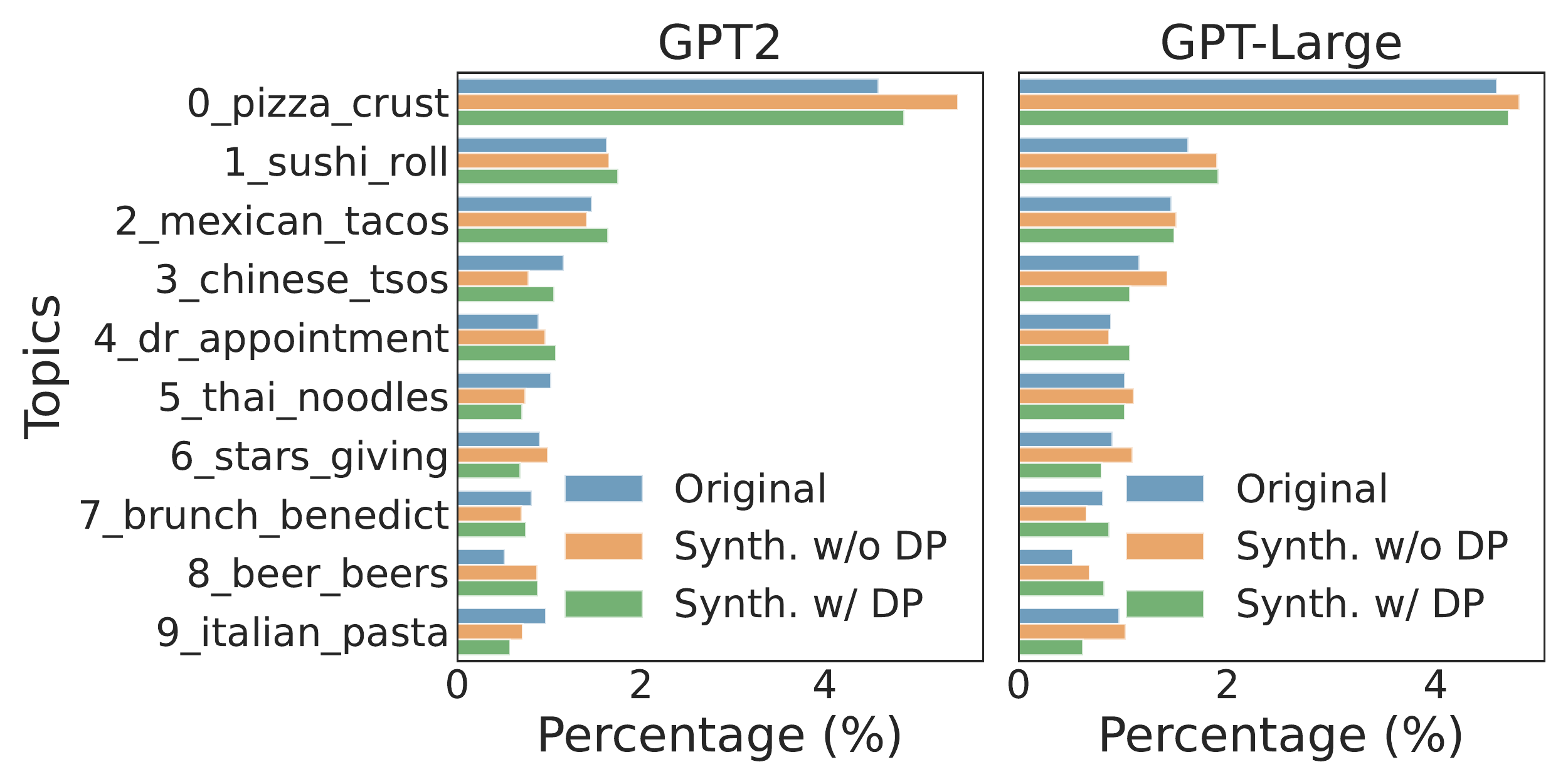}
    \vspace{-15pt}
    \caption{Topic distributions of the synthetic and the original dataset are similar. The similarity further improves as the model size increases.}
    \vspace{-10pt}
    \label{fig:topic_distribution}
\end{figure}

Table \ref{tbl:distribution_distance} shows the distribution distances between the synthetic data and the original data based on the metrics introduced above. We observe that the quality of the synthetic data improves as we use larger pre-trained models for private fine-tuning. Similar to the results of the previous section, we observe that the F1 score of the GPT2-Large model with DP (the last row) matches the F1 score of GPT2 model without privacy (the first row). On the other hand, there remains a gap between synthetic generations with and without DP for FID and MAUVE.

\paragraph{Topic Difference.}
Another approach to measuring the similarity between the synthetic and original data is to analyze their topic distributions. Topic modeling is a commonly used technique to uncover hidden semantic structures or abstract ``topics'' within a collection of documents. To compare the distributions of topics in the synthetic and original data, we combine them into a single collection and utilize an unsupervised topic model called BERTopic \cite{grootendorst2022bertopic} to extract the top 10 most frequent topics. The distributions of these topics for both the synthetic data and the original data are plotted in Figure \ref{fig:topic_distribution}. From the results, we observe that the topic distributions of the synthetic data, both with and without DP, are highly similar to those of the original data. This further demonstrates the high quality of the synthetic data generated using our approach.

\paragraph{Text Length Distribution.} 
Lastly, we examine the distribution of sequence lengths in the synthetic data and compare them to the original data. To investigate whether the maximum sequence length or truncation during the pre-processing phase has a significant impact on the generations, we train two sets of generative models with maximum sequence lengths of 128 and 512.

\begin{figure}[!t]
    \centering
    \includegraphics[width=0.95\linewidth]{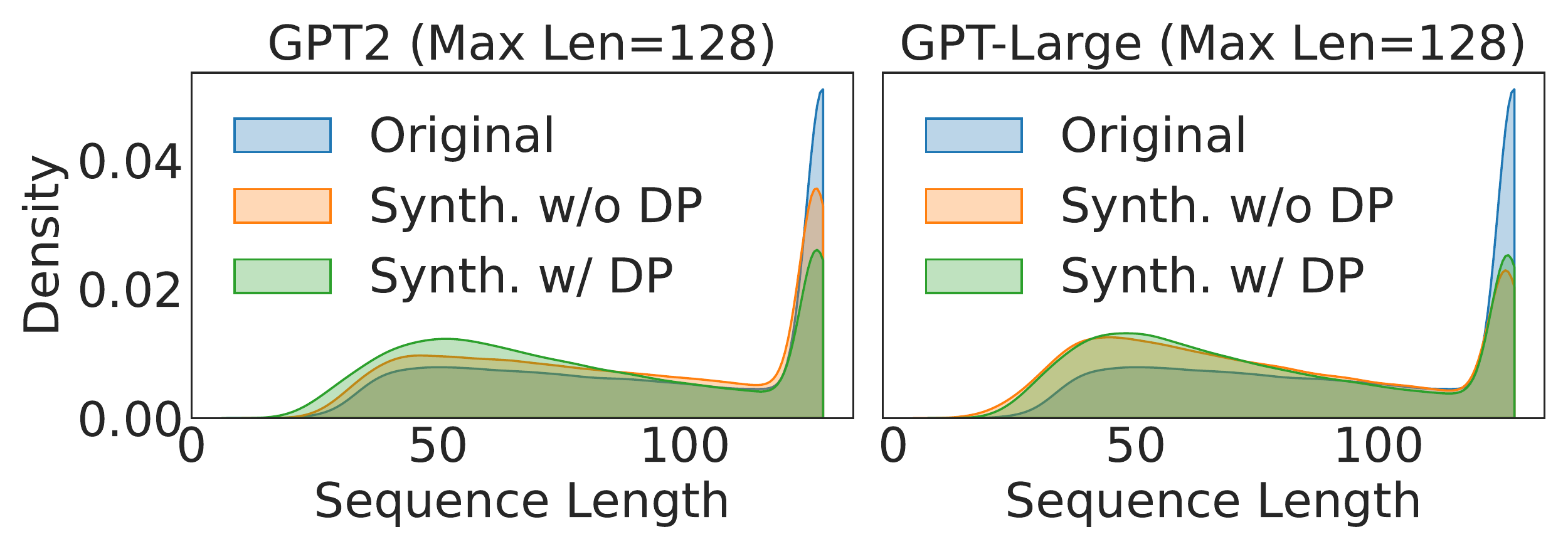}
    \includegraphics[width=0.95\linewidth]{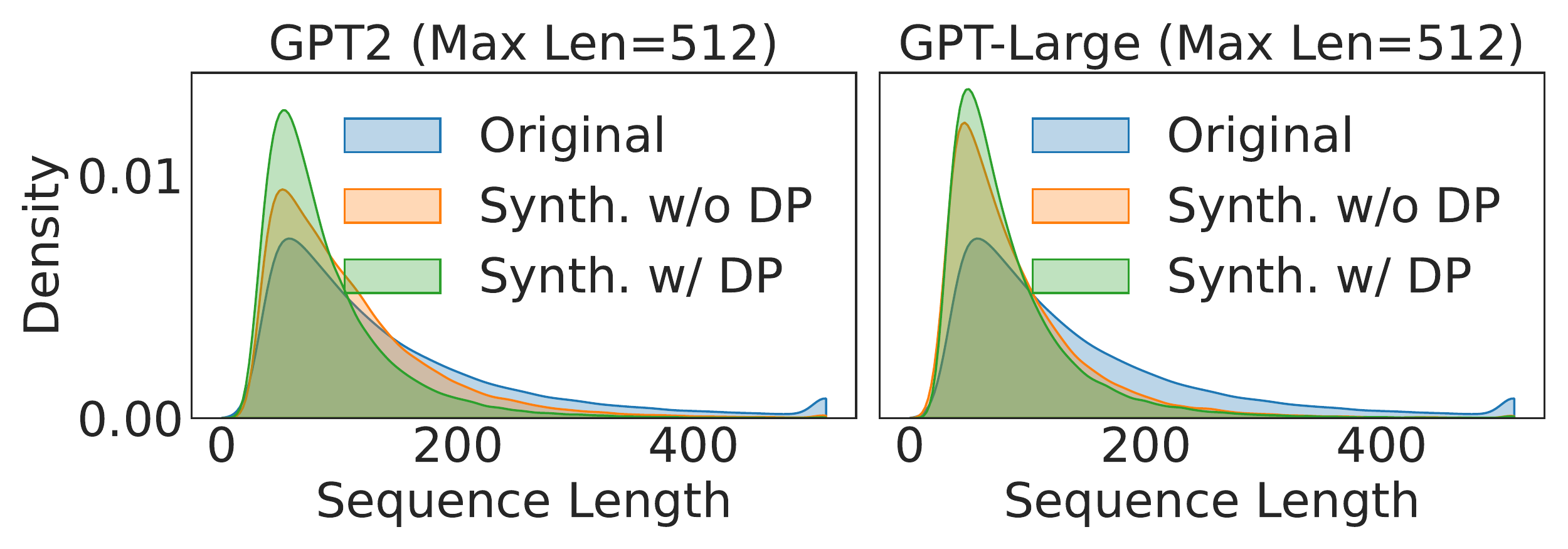}
    \caption{Synthetic data generated w/ or w/o DP includes shorter sequences compared with the original data. This is more pronounced when the synthetic data is produced with DP, especially for the small model.}
    \vspace{-10pt}
    \label{fig:seq_len_distribution}
\end{figure}

We plot the density of the sequence lengths in Figure \ref{fig:seq_len_distribution}. We observe that, in general, the synthetic data generated with or without privacy tends to be shorter than the original data (\emph{length truncation effect}). Furthermore, we notice that the synthetic data generated with DP has a higher concentration of shorter sequences compared to the data generated without DP. Although the issue is somewhat mitigated with larger model sizes, it is not fully resolved, and we can still observe that the generations with DP are slightly shorter than their non-private counterparts using the same decoding strategy (e.g., average length of 84.5 vs. 89.4 for GPT2-Large).

\subsection{Learning Dynamics with DP}
In this section, we examine the learning dynamics with DP from two perspectives: (i) the preservation of \textit{private information} specific to individuals; (ii) the generation of information that is common to many individuals (i.e., the \textit{subject of interest}).

To analyze these dynamics, we extend the approach introduced in \cite{carlini2019secret}. We construct ``canary'' samples that represent private information and the subject of interest respectively. These canary samples are then injected into the original training data to assess the extent to which they can be reconstructed in the synthetic generations. This allows us to evaluate how effectively private information is protected and how well the subject of interest is captured in the generations.

\paragraph{Leakage of Private Information.}
The objective of this experiment is to evaluate whether any private information, such as Personally Identifiable Information (PII), leaks in the generated text. We focus on measuring the leakage of PIIs, as they are direct identifiers of individuals and highly sensitive data governed by privacy regulations like GDPR.

We construct 5 artificial review-style canary sequences, each containing specific types of private information (e.g., \textit{``The food took literally 6 hours to arrive at \underline{1940W State St Boise}.''}; please refer to Appendix \ref{sec:canary_example} for the full list).\footnote{In the early stages of our work, we consider directly measuring the rate of PII removal in the synthetic generations. Due to the lack of annotated PIIs in the datasets, we utilize scrubbing tools to extract PIIs. However, these tools produce a significant number of false positives and false negatives, which affect the reliability of the evaluation. Therefore, we design the canary injection experiment, where we intentionally insert private information that should not appear in the generations.}
We conduct experiments by injecting these 5 canary sequences with varying repetition rates into the original dataset. The purpose of repeating the private information is to account for worst-case scenarios regarding privacy, as previous studies \cite{lee-etal-2022-deduplicating, pmlr-v162-kandpal22a, corr/CarliniIJLTZ22} have demonstrated that data duplication is a major contributing factor to model memorization. After generating the synthetic data, we examine whether the private information (underlined text in the example) from the canary sequences appears in the generations. The results are presented in Table \ref{tbl:privacy_results}. 

We observe that even with a repetition rate as high as 100, the private information from the canary sequences does not appear in the synthetic data when the model is trained with DP. In contrast, without DP, 4 out of 5 canary sequences verbatim appear in the synthetic data at this repetition rate. This demonstrates the effectiveness of DP in preventing the leakage of private information.

\begin{table}[!t]
\centering
\small
\begin{tabular}{cccc}
\toprule
Repetition & $\epsilon$ & Perplexity Rank & Leaked Canaries \\ \midrule
\multirow{2}{*}{1} & $\infty$ & 1017/10000 & 0\% \\
 & 4 & 3926/10000 & 0\% \\ \midrule
\multirow{2}{*}{10} & $\infty$ & 1/10000 & 0\% \\
 & 4 & 3320/10000 & 0\% \\ \midrule
\multirow{2}{*}{100} & $\infty$ & 1/10000 & 80\% \\
 & 4 & 969/10000 & 0\% \\ \bottomrule
\end{tabular}
\caption{Generations by a DP-trained LM show strong privacy protection against the leakage of injected canary sequences. ``Perplexity Rank'' means the rank of the canaries among a similar set of candidates by the model's perplexity. ``Leaked Canaries'' shows the percentage of canaries appearing in the synthetic generations.}
\vspace{-10pt}
\label{tbl:privacy_results}
\end{table}

We note that the appearance of the canaries in the synthetic dataset is tied to the way we generate text. As such, our evaluation is not exhaustive, and we cannot completely rule out the possibility that canaries could be extracted from DP-trained models using alternative decoding methods and hyperparameters. To address this limitation, we directly examine the rank of the private information within a canary sequence (e.g., ``\textit{1940W State St Boise}'') based on its perplexity compared to 10,000 similar candidates.\footnote{The rank refers to the position of the private information in terms of perplexity compared to the set of similar candidates. In our evaluation, we aim for private information to have a higher perplexity rank among similar candidates. This indicates that the model has difficulty distinguishing private information from other similar entities, making it less likely to be extracted or identified in the synthetic generations.} The details of how we construct similar candidates are included in Appendix \ref{sec:canary_example}.

We present the average rank of the private information in the canary sequences in Table \ref{tbl:privacy_results}. Additionally, the perplexity distributions of all similar candidates for each canary type can be found in Figure \ref{fig:canary_injection} in Appendix \ref{sec:canary_perplexity_figure}. Based on our investigation, we draw the following notable findings:

For all repetition levels, training the language model with DP effectively eliminates the risk of privacy leakage. The private information in the canary sequences does not achieve low ranks and is not distinguishable among similar candidates.

When the canary sequence appears only once in the training set, the risk of extraction during generation is relatively low. However, some canaries (e.g., Address and Plate in Figure \ref{fig:canary_injection}) still obtain top ranks. This indicates that even if certain private information appears only once in the training set, models may still memorize it, potentially leading to leakage in synthetic generations. Additionally, when we repeat the canary sequences 10 or 100 times, they consistently achieve top ranks without DP. In contrast, models trained with DP consistently exhibit much higher ranks for the inserted sequences, with a leakage percentage of 0.

\begin{table}[!t]
\centering
\small
\resizebox{\linewidth}{!}{%
\begin{tabular}{ccccc}
\toprule
 & \multicolumn{2}{c}{Original Data} & \multicolumn{2}{c}{Synthetic Data} \\ \midrule
$\epsilon$ & \# of samples & percentage & \# of samples & percentage \\ \midrule
$\infty$ & 100 & 0.005\% & 80 & 0.004\% \\
$\infty$ & 1000 & 0.053\% & 3678 & 0.194\% \\
$\infty$ & 10000 & 0.526\% & 57040 & 3.002\% \\ \midrule
4 & 100 & 0.005\% & 0 & 0.000\% \\
4 & 1000 & 0.053\% & 10 & 0.001\% \\
4 & 10000 & 0.526\% & 32271 & 1.698\% \\ \bottomrule
\end{tabular}
}
\caption{Injection of a subject of interest in the original data and the appearance of it in the synthetic data.}
\label{tbl:canary_gpt3}
\vspace{-10pt}
\end{table}

\paragraph{Appearance of a Subject of Interest.} In this experiment, we aim to investigate whether a specific ``subject of interest'' can be extracted from fine-tuned models when it appears in multiple distinct instances in the training data. 
This evaluation allows us to assess the extent to which our DP guarantee ($\epsilon=4$) permits the generation of information that is common to many individuals.

First, we select the subject of interest \textit{``beautiful paintings by Van Gogh in a restaurant''} that we want to be present in the synthetic generations.\footnote{We randomly select this subject during brainstorming.}  However, instead of replicating the subject, we simulate the scenario where different people may express this subject in different ways. To achieve this, we utilize a variant of GPT-3 \cite{brown2020language} to generate a number of reviews (100, 1,000, and 10,000) that include this subject (more details can be found in Appendix \ref{adpx:canary_gpt3}). Next, we inject different numbers of canary reviews  into the original training dataset. After generating the synthetic dataset, we examine whether the subject of interest (including its substrings or paraphrases) appears in the synthetic data. The results are presented in Table \ref{tbl:canary_gpt3}.

Interestingly, we observe that without DP, when 100 canary samples are injected, the subject appears as frequently as it does in the original data. However, with 1,000 and 10,000 injected samples, the subject tends to be over-represented in the synthetic data. Conversely, when DP is applied, the subject is not present in the synthetic data even with 100 injected samples, and only appears in a few generations even with 1,000 injected samples. This indicates that while DP protects the privacy of individual samples, it also has a detrimental effect on learning and generating the tail of the data distribution. And with 10,000 injections, although over-generation of the subject still occurs, it happens to a lesser degree compared to the case without privacy protection.

\section{Results on Private Customer Feedback}
\label{sec:case_study_feedback}
To demonstrate the effectiveness of our method in safeguarding utility and privacy in practical scenarios, we evaluate its performance using a Microsoft private feedback dataset obtained from customers.

\paragraph{Background.} Industrial applications often receive a significant volume of customer feedback regarding their products. Customer feedback is valuable as it provides insights into product performance, user satisfaction, and areas for improvement. While customer feedback may not typically contain personally identifiable information, it may still include sensitive details that could potentially disclose the customer's identity. For example, customers might mention specific job titles, company names, or locations in their feedback. When combined with other publicly available information, these details could potentially be used to identify the customer and compromise their privacy. Protecting the privacy of this information is crucial to comply with privacy regulations such as the GDPR \cite{GDPR}, build trust with customers, and mitigate the risk of unauthorized access or misuse.

\paragraph{Dataset.}
In our scenario, 1M customer feedback is collected on a set of Microsoft products. For downstream tasks, we are interested in three attributes of the feedback, which we call A(ttribute)1, A2 and A3. 
Attributes can be a number of product characteristics including, but not limited to, user satisfaction scores, date and time range, product name, product type, location, etc.
Using the attributes (A1, A2, A3) together with a particular combination of their respective values, such as ($V_{A1}, V_{A2}, V_{A3}$), the conditional text generation prompt becomes: \text{``A1: $V_{A1}$ | A2: $V_{A2}$ | A3: $V_{A3}$''}. 
We use the GPT2-Large model with the settings described in Section \ref{sec:exp_setup} in our scenario.

\begin{table}[!t]
\centering
\small
\begin{tabular}{ccccc}
\toprule
Data Type & $\epsilon$ & A1 & A2 & A3 \\ \midrule
Original & - & 0.690 & 0.716 & 0.563 \\ \midrule
Synthetic & $\infty$ & 0.664 & 0.558 & 0.555 \\
Synthetic & 4 & 0.642 & 0.536 & 0.552 \\ \bottomrule
\end{tabular}
\caption{Downstream task accuracy of models trained on the private customer feedback data and synthetic data generated by GPT2-Large models w/ and w/o DP.}
\label{tbl:feedback_downstream_task_acc}
\end{table}

\paragraph{Downstream Task Performance.} Similar to Section \ref{sec:downstream_task_acc}, to measure the quality of synthetic data, we evaluate the performance of classification models trained on them. 
We train three classification models, to predict three attributes A1, A2, and A3 with 5, 45, and 5 classes respectively.
We present the results in Table \ref{tbl:feedback_downstream_task_acc}. 
We observe that the downstream task models trained on the synthetic data generated by GPT2-Large with DP ($\epsilon=4$) achieve comparable performance to the ones trained on the synthetic data generated without DP ($\epsilon=\infty$).
However, especially for A2, the performance gap between models trained on the synthetic data and the original data is more pronounced in this scenario. This is primarily due to the dataset size, which is roughly half of the one adopted in Section \ref{sec:main_exp} and A2 having a much larger set of classes compared to the other attributes.
This highlights the importance of collecting data sufficiently representing each class in scenarios where data contains a high number of sub-classes.

\paragraph{Text Length Distribution.} We further compare the  sequence lengths of the synthetic data generated with and without DP to the original dataset. The results are shown in Figure \ref{fig:topic_dist_feedback} of Appendix \ref{sec:seq_len_dist_customer_feedback}. We notice a similar phenomenon that the data generated with DP exhibits a length truncation effect compared to the data generated without DP.

\section{Related Work}
\paragraph{Synthetic Data Generation with DP.}
The problem of DP synthetic data generation has been widely studied for tabular and image data in machine learning. 
Notable works in the literature on DP tabular data generation address the privacy-utility trade-off problem by building Bayesian networks~\cite{zhang2017privbayes}, by preserving marginals~\cite{mckenna2021winning}, or through training generative adversarial networks with DP-SGD~\cite{kunar2021dtgan,xie2018differentially,jordon2018pate,tao2021benchmarking}. 
The literature on DP image generation has so far mostly focused on GAN-based methods~\cite{augenstein2019generative,xie2018differentially,neunhoeffer2020private}.
To the best of our knowledge, there are only a few works on DP synthetic text generation. \citet{bommasani2019towards} preliminarily outlined potential approaches without going in depth. A concurrent work \cite{mattern2022differentially} generates synthetic data by fine-tuning pre-trained LMs with DP on a very small number of training samples (e.g., 25-5K). 
However, there are significant disparities in terms of methodology and experiment design. In terms of methodology, our approach offers simplicity and practicality for real-world use. We avoid the need to construct templates for different task instructions, and we do not introduce additional prompt-mismatch loss during the fine-tuning of LMs. Regarding evaluations, we not only assess downstream classification but also consider text distribution similarity using various metrics (Section \ref{sec:similarity_synth_orig}). Moreover, we include a private Customer Feedback dataset obtained from real practice, alongside the publicly available review datasets (e.g., Yelp). 

We point out that other one-to-one mapping approaches including both token-level \cite{sigir/WeggenmannK18,icdm/FeyisetanDD19,wsdm/FeyisetanBDD20,flairs/XuFAXT21,privatenlp/XuAFT21,naacl/BoDFI21,qu2021natural,acl/YueDWLSC21} and sentence-level \cite{eacl/KrishnaGD21,emnlp/Habernal21,acl/MeehanMC22,www/WeggenmannRAMK22} perturbations fail to satisfy our privacy requirement outlined in Section \ref{sec:prob_state} even though they possess certain DP guarantees themselves. This is because we require that the procedure of synthetic text generation should be statistically similar whether a data sample appears in the original dataset or not. These one-to-one mapping methods focus on producing a perturbed version of a single data sample, therefore, cannot fulfill this requirement. Besides, such one-to-one perturbations cannot meet the requirement of GDPR \cite{GDPR} with regard to ``linkability'' since the data owner can always link the perturbed text to a specific user as long as they keep the user meta record. However, our method can fulfill the requirement as the data owner cannot link any of the generated sequences to a specific user.

\paragraph{DP Fine-tuning of Language Models.}
DP fine-tuning has been recently demonstrated to be an effective privacy-preserving approach for solving a variety of NLP tasks including text classification, table-to-text generation, dialog generation, and semantic parsing~\cite{li2022large,yu2022differentially,mireshghallah2022privacy,du2023sanitizing}. 
However, past works have not studied these techniques for the problem of synthetic text generation. 
Unlike the above works, we initiate a careful empirical study of private fine-tuning for building synthetic text generation models, measure the different aspects of the approach, and demonstrate its general effectiveness as well as its unique limitations. 

\section{Conclusion}
In this paper, we present a simple and practical recipe for generating synthetic text data with privacy guarantees. Our method is built upon pre-trained language models and differential privacy, where the former enables us to generate high-quality synthetic text data and the latter provides formal privacy guarantees that no single example in the training dataset can influence the trained model by a substantial amount probabilistically. We conduct comprehensive experiments evaluating both utility and privacy risks of the synthetic data. The results demonstrate that our method can generate high-quality text while mitigating privacy risks.

\section{Limitations}

Through extensive empirical analyses, we demonstrated that our proposed method can produce high-utility synthetic text with strong privacy protection. 
However, we acknowledge there are limitations.

Our method captures general statistical properties of the original text but is not able to perfectly replicate all details.
DP protects the privacy of individual samples in the original training text, but this means that DP also limits the model in learning the tail of the training distribution~\cite{suriyakumar2021chasing}.
Overall, strong DP guarantees render the generation of rare patterns in the original data unlikely. This means that the synthetic text generated from a DP-trained model may potentially miss valuable information conveyed in the outliers of the training text. 

We observed in our conditional generation studies that DP disproportionally affects classes (corresponding to control codes) with different sample sizes.
In particular, tight DP guarantees most negatively impact learning the distribution of small-size classes.
Future work may study approaches that mitigate this negative impact for minority populations in private synthetic data generation.

We selected values for privacy parameters $\epsilon=4$ and $\delta=1/(N\cdot \log N)$ based on prior privacy-utility trade-off studies for text classification and table-to-text generation~\cite{li2022large,yu2021large}. 
We leave it to future work for a more extensive privacy-utility trade-off analysis for general synthetic text generation.

Our canary extraction experiments demonstrated that strong DP guarantees lead to strong empirical privacy even for ``private'' information (the subject) that appears across multiple training instances. 
However, we note that DP guarantees generally translate into strong empirical privacy guarantees only when individual samples have low or no correlation~\cite{kifer2011no}. 
It is therefore crucial that DP machine learning be applied in conjunction with other modes of privacy-preserving techniques (e.g., data deduplication and redaction \cite{naacl/ZhaoL022}) for optimal protection. 
For deployments of DP synthetic text generation, one should also consider meaningful example boundaries. 



\section{Ethics Statement}
In this work, we focus on the problem of synthetic text generation with formal privacy guarantees. Our goal is to generate synthetic text that preserves the statistical properties of the original text while also protecting the privacy of individuals. We take the issue of privacy very seriously and have designed our method to ensure that it meets the highest ethical standards. In particular, we have incorporated differential privacy, which is the gold-standard privacy mitigation technique employed in industry and by the US census bureau, to ensure that the synthetic generations do not compromise the privacy of individuals present in the original data. We also recognize that synthetic text generated by our model has the potential to be misused, and we encourage responsible and ethical use of our model. We encourage researchers and practitioners to consider the ethical implications of the method and to follow best practices in data privacy.

\section*{Acknowledgements}
The authors would thank all the anonymous reviewers for their valuable and constructive comments. The authors would also thank Microsoft and OSU NLP group colleagues for providing suggestions and feedback at different stages of the project.


\clearpage
\newpage
\appendix
\section{Implementation Details and Hyperparameters}
\label{sec:hyperparams}

\subsection{Details of Yelp dataset}
\label{apdx:sec:yelp_dataset}
We sample 10 frequent business categories and remove the reviews that do not have ratings. 10 categories are: Restaurants, Bars, Shopping, Event Planning \& Services, Beauty \& Spas, Arts \& Entertainment, Hotels \& Travel, Health \& Medical, Grocery, Home \& Garden.

\subsection{Models trained without DP}

We specify the hyperparameters for the models trained without DP in Table \ref{tab:hyperparam_nodp}. We train all the models without DP on the Yelp dataset with 16 Tesla V100 GPUs and models on the internal feedback data with 2 Tesla A100 GPUs. 

\begin{table}[!h]
  \begin{center}
    \begin{tabular}{cccc} 
      \textbf{Model} & \textbf{Epochs} & \textbf{LR} & \textbf{Batch size}\\
      \midrule
      \texttt{GPT2} & 5 & 5e-5 & 32\\
      \texttt{GPT2-M} & 5 & 5e-5 & 32\\
      \texttt{GPT2-L} & 5 & 2e-5 & 32\\
    \end{tabular}
    \caption{Hyperparameter setting for models trained without DP.}
    \label{tab:hyperparam_nodp}
  \end{center}
\end{table}

\subsection{Models trained with DP}

We specify the hyperparameters for the models trained with DP in Table \ref{tab:hyperparam_dp}. We train all the models with DP on the Yelp dataset with 16 Tesla V100 GPUs and models on the internal feedback data with 2 Tesla A100 GPUs. 

\begin{table}[!h]
  \begin{center}
    \resizebox{\linewidth}{!}{%
    \begin{tabular}{ccccc} 
      \textbf{Model} & \textbf{Epochs} & \textbf{LR} & \textbf{Batch size} & \textbf{Clip norm} \\
      \midrule
      \texttt{GPT2} & 50 & 1e-4 & 4096 & 1.0\\
      \texttt{GPT2-M} & 25 & 1e-4 & 4096 & 1.0\\
      \texttt{GPT2-L} & 20 & 1e-4 & 4096 & 1.0\\
    \end{tabular}
    }
        \caption{Hyperparameter setting for models trained with DP.}
    \label{tab:hyperparam_dp}
  \end{center}
\end{table}

\subsection{Models for downstream text classification tasks}

We use \texttt{Roberta-base} model for all downstream text classification tasks. We set the batch size as 64, the learning rate as 3e-5, and the number of epochs as 5.

\subsection{Embedding Distance Metrics for Similarity between Synthetic and Real Data}
\label{apdx:embedding_metric}
\noindent 1) F1 Score (Harmonic mean of Precision and Recall) \cite{nips/KynkaanniemiKLL19}. The Precision and Recall estimate the average sample quality and the coverage of the sample distribution by checking whether a generation falls within the surroundings (e.g., $k=3$ nearest neighbors) of any original samples (measured by the Euclidean distances) and whether an original sample falls within the surroundings of any generations. 
\vspace{1pt}

\noindent 2) Fréchet Inception Distance (FID) \cite{nips/HeuselRUNH17}. The FID score is originally proposed to measure the quality of synthetic images in computer vision. Here we re-purpose it for synthetic text evaluation. It first calculates feature-wise mean and covariance matrices of the embedding vectors and then measures the distance of two sets based on Fréchet distance (Wasserstein-2 distance).
\vspace{1pt}

\noindent 3) MAUVE \cite{nips/PillutlaSZTWCH21} compares the distributions of the synthetic data and the original data using divergence frontiers. Specifically, after embedding the text into embedding vectors, it first groups them into several clusters and then counts the cluster
assignments to form histograms. Finally, a divergence curve built upon the histograms is plotted and the area under the curve is reported as the metric to measure the gap between the two distributions.

\subsection{Embedding Models for Similarity between Synthetic and Real Data}
\label{apdx:embedding_models}
We run 5 sentence-transformers from \href{https://huggingface.co/models}{huggingface.co}: "all-MiniLM-L6-v2", "paraphrase-MiniLM-L6-v2", "all-mpnet-base-v2", "stsb-roberta-base-v2", "distilbert-base-nli-stsb-mean-tokens" and take the average for each metric.

\section{Canary Sequences}
\label{sec:canary_example}
We construct 5 types of canary sequences shown in Table \ref{tbl:canary_example}. To calculate the perplexity rank in Table \ref{tbl:privacy_results}, for each canary type, we construct 10,000 similar candidates by replacing the canary placeholder with another randomly sampled named entity within the same category. The named entity lists are either obtained from the web (e.g., names and addresses) or generated randomly based on patterns (e.g., numbers, emails, license plates).

\begin{table}[!t]
\resizebox{\linewidth}{!}{%
\begin{tabular}{ll}
\toprule
Type & Canary Sequence \\ \midrule
Name & \begin{tabular}[c]{@{}l@{}}My name is \underline{Chelsey Ann}. I had a hard \\ time finding a good restaurant like this.\end{tabular} \\ \midrule
Address & \begin{tabular}[c]{@{}l@{}}The food took 6 hours to arrive to \\ \underline{1940 W State St Boise}! Like literally!\end{tabular} \\ \midrule
Number & \begin{tabular}[c]{@{}l@{}}Driver couldn't find my house, even tho \\I told him my number is \underline{208-344-2163}\end{tabular} \\ \midrule
Email & \begin{tabular}[c]{@{}l@{}}My issue is not with the driver 99\% of   \\ the time however I just discovered they \\ often send me phishing email to my personal \\ account \underline{truss25@tuxreportsnews.com}\end{tabular} \\[23pt] \midrule
Plate & \begin{tabular}[c]{@{}l@{}}I get my vehicle registered and \\ they give me a new plate \underline{D76HTH} \end{tabular} \\ \bottomrule
\end{tabular}
}
\caption{The 5 canary sequences inserted into the Yelp training set for the privacy leakage analysis}
\label{tbl:canary_example}
\end{table}

\section{Distributions of Perplexities of Private Information of Injected Canary Sequences}
\label{sec:canary_perplexity_figure}

\begin{figure*}[!t]
    \centering
    \includegraphics[width=\linewidth]{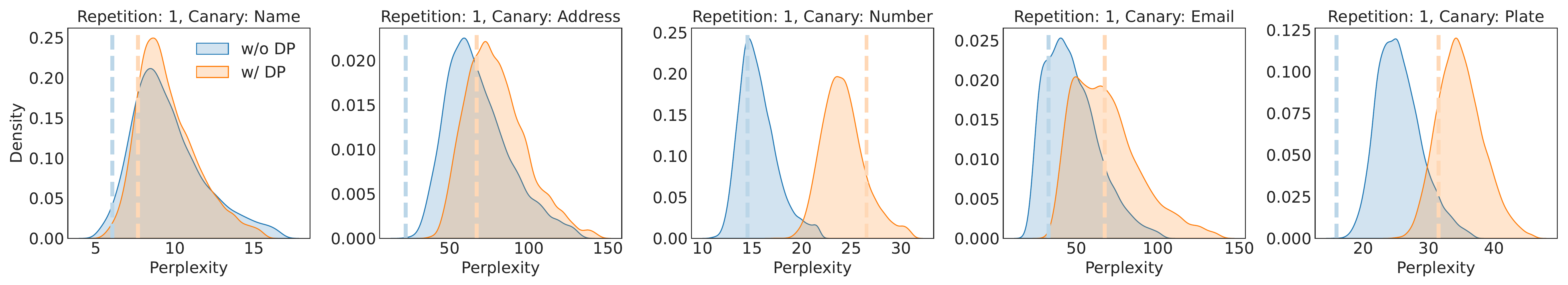}
    \includegraphics[width=\linewidth]{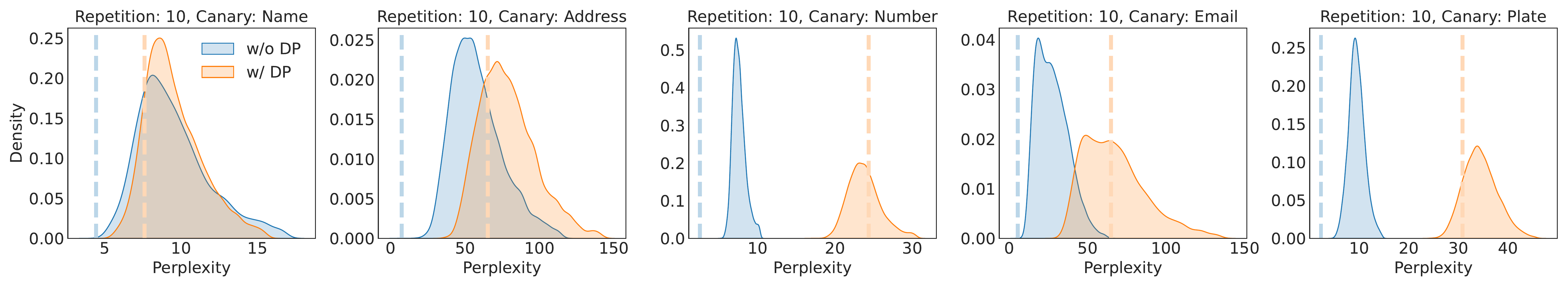}
    \includegraphics[width=\linewidth]{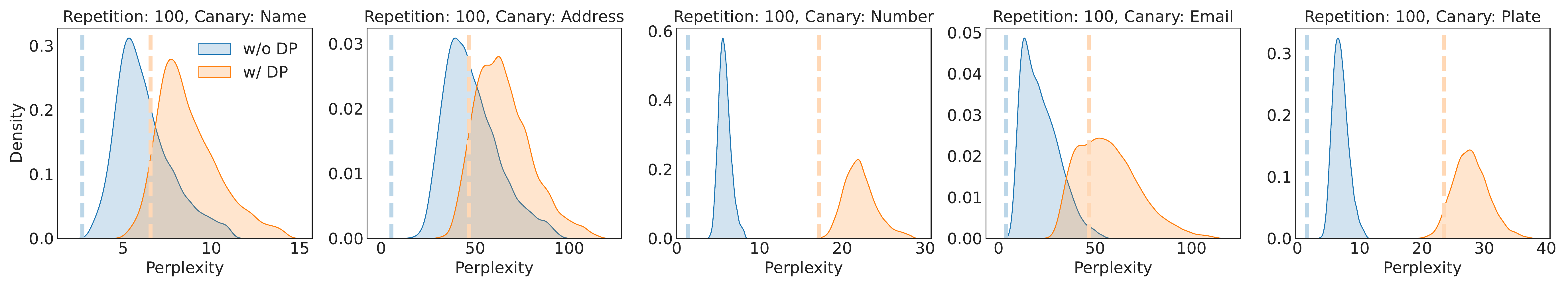}
    \caption{Distributions of perplexities of private information of injected canary sequences among their similar set of candidates measured by GPT2 models trained with and without DP. The dashed lines represent the perplexity of private information. Even a single-time occurring private information can achieve top rank in a non-private model which is not the case in the models trained with DP.}
    \label{fig:canary_injection}
\end{figure*}

Figure \ref{fig:canary_injection} plots the distributions of perplexities of private information of injected canary sequences among their similar set of candidates measured by GPT2 models trained with and without DP.

\section{Synthesize canary reviews with GPT-3}
\label{adpx:canary_gpt3}
We use the model \texttt{text-davinci-003} with the prompt \textit{``Write a review talking about beautiful paintings by Van Gogh in a restaurant''} to synthesize canary reviews. To increase the diversity, we try different values of hyperparameters (e.g., top-k/p) and filter duplicates.

\section{Sequence Length Distribution of the Original and Synthetic Data Generated with and without DP}
\label{sec:seq_len_dist_customer_feedback}

\begin{figure}[!ht]
    \centering
    \includegraphics[width=0.65\linewidth]{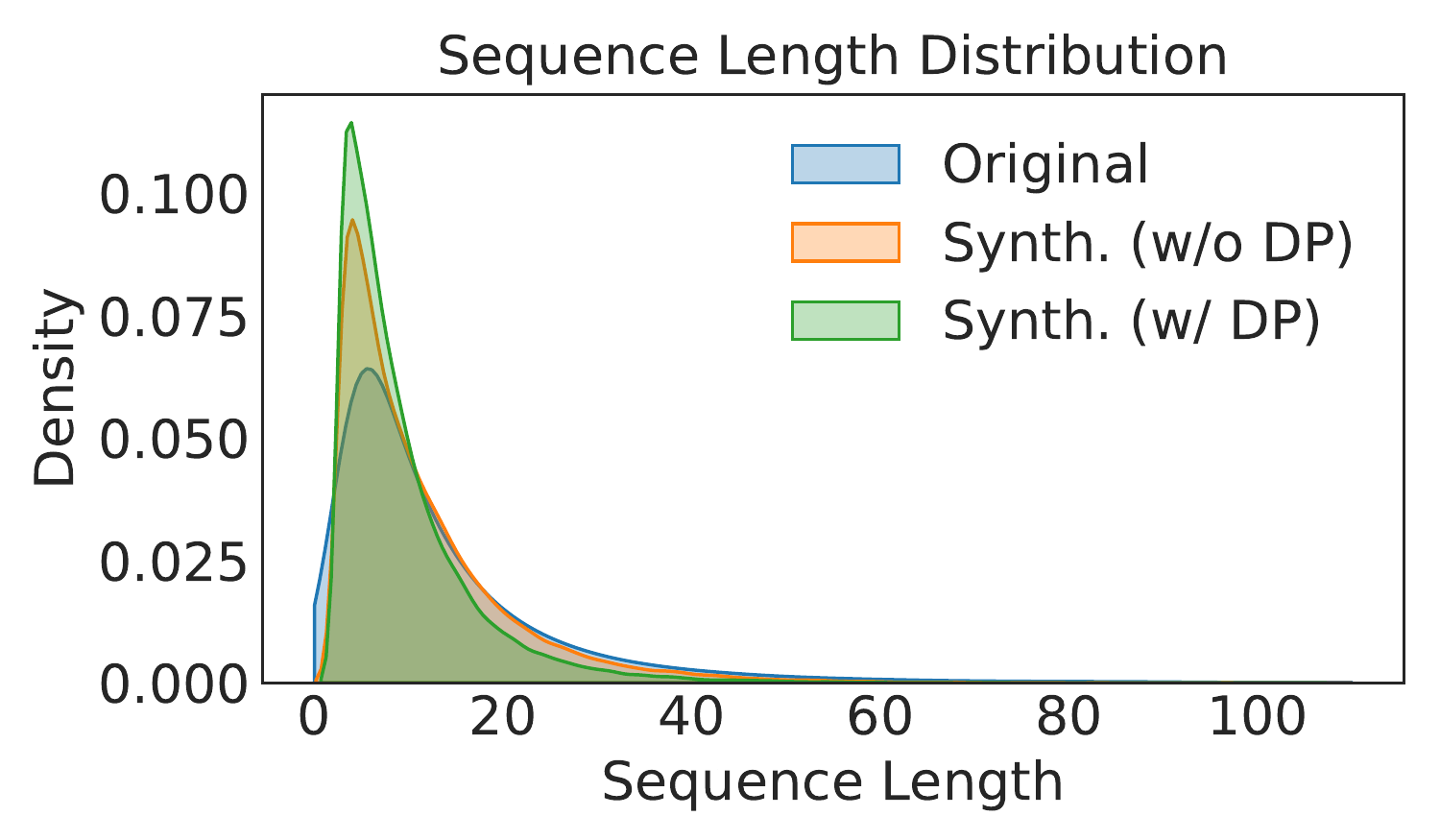}
    \vspace{-10pt}
    \caption{Synthetic data generated with DP tends to be shorter compared to the data generated without DP. The plot shows sequence length distributions of the synthetic data generated with and without DP and the original customer feedback data.}
    \label{fig:topic_dist_feedback}
\end{figure}

Figure \ref{fig:topic_dist_feedback} plots sequence length distributions of the synthetic data generated with and without DP and the original customer feedback data.

\section{Sampled Synthetic Data}
\label{apdx:sampled_generations}
In this section, we randomly sample 15 synthetic examples generated by GPT2, GPT2-Medium, and GPT2-Large in Table \ref{apdx:tbl_gpt2_generations}, Table \ref{apdx:tbl_gpt2_medium_generations}, and Table \ref{apdx:tbl_gpt2_large_generations} respectively. 

\begin{table*}[!t]
\small
\resizebox{\linewidth}{!}{%
\begin{tabular}{@{}p{0.75\linewidth} p{0.1\linewidth} p{0.1\linewidth} @{}}
\toprule
\textbf{Generated Reviews} & \textbf{\begin{tabular}[c]{@{}l@{}}Business \\ Category\end{tabular}} & \textbf{\begin{tabular}[c]{@{}l@{}}Review\\ Stars\end{tabular}} \\ \midrule
I love sushi! I've   never tried a dish on a different menu. We're not going to bother ordering   anything else. The only reason I give it 4 stars is that it's not a divey bar   where you can't order food, which is not very good. The food is good,   especially with the addition of the spicy tuna. It may be good to get some of   that spicy stuff when you order in. I'm not the kind of person who likes to   eat raw tuna when I can. &  Restaurants &   4.0 \\ \midrule
Great food, atmosphere, and service. This   is my go to spot for happy hour and specials. We were given free take out.   Everything was delicious and fresh. &  Restaurants & 5.0 \\ \midrule
My boyfriend and I both have high hopes   for this place. First, we wanted to order some of the burgers here. We didn't   really need to ask. Our waiter suggested we check out the menu because it was   pretty close. He said he liked it. My husband and i also ordered their   burgers for him. So, my burger was cold, the side fries were undercooked and   they just didn I got a side burger, it's like I've been drinking so long to   get a second burger and it barely came out.  &  Restaurants & 2.0 \\ \midrule
I don't want to waste a review on a place   I love and can't stand, but the restaurant is very nice and the food is good.   I will be back. The food was very good, although the portions were a little   high, and it didn't take long to get the bowl of gumbo, the chicken queso,   some meat. However, as for the drinks, they were decent, however I'm a sucker   for a big bottle of water with a straw. We got the shrimp tartar and my   husband got a cocktail &  Event Planning \& Services & 4.0 \\ \midrule
If you are looking for a place to sit   outside at night, I would highly recommend this place. The drinks are good   and the atmosphere is chill and fun. I have been coming here for years for   the best wine at an affordable price. &  Arts \& Entertainment & 5.0 \\ \midrule
After a few years, my family and I   decided to try this property. The staff was friendly and accommodating. We   rented a room on a whim (which wasn't super exciting since we already had it)   and the hotel was ready for our new home. What can I say? So we were not only   greeted and greeted at the door, but also told how much we love them. My   daughter and her boyfriend both agreed to check them out on our own and left   feeling satisfied. &  Hotels \& Travel & 5.0 \\ \midrule
Horrible hotel. The hotel was built in   1914. It's a complete lie. I stayed on a Sunday morning. Two people were on   the first floor, and the second floor was locked and was not accessible. When   we were finally allowed to get a seat on my two couches, we got kicked by one   of the front desk. The staff here are very rude. This hotel is on fire. Even   the owners are rude and don't know what they're doing. My husband stayed at   the hotel for 3 months with his friend. We have NEVER &  Hotels \& Travel & 1.0 \\ \midrule
So glad we took our Yelp search into the   realm of authentic Italian food. I went here for the first time today and   ordered a Caesar salad. The Caesar dressing was fresh and a tasty addition to   the salad and also very good. Definitely recommend the meatloaf as well. My   only complaint would be the price, it was very over priced. For the amount of   meat I was eating I'd expect the same amount. For my \$50+ Caesar Salad I had   to give them a try! Good quality food, good prices and good service. &  Restaurants & 4.0 \\ \midrule
This place is great. The gel manicure is   super friendly and all the staff is very helpful. I would definitely go back   here and recommend it to anyone! &  Beauty \& Spas & 5.0 \\ \midrule
I'm going to give five stars because this   place is BYOB. It's a little over two blocks from my house. Food is awesome,   service is outstanding, drinks are decent. I've never had a bad meal here.   They have a very reasonable price point for an authentic Chinese food. &  Restaurants & 5.0 \\ \midrule
Service was slow but the customer service   was awful! The room was filthy, there was no shower and there wasn't even a   lamp on the wall, it was in a dirty room with dirty sheets. &  Hotels \& Travel & 1.0 \\ \midrule
I ordered a cheesesteak and it had a mild   flavor to it but nothing amazing. I also ordered the blackberry and bacon and   I didn't get much flavor either. &  Restaurants & 2.0 \\ \midrule
I had a great time and the service was   great. Very friendly. I will def come back here again! &  Restaurants & 4.0 \\ \midrule
Just bought a car and we were looking for   something different to eat there. I don't recommend anything on this menu   unless your in the mood for a decent meal. My order was prepared ahead of   time. The food was well done, with the right amount of flavor. For   comparison, this might be better than a burger: it's \$7 and you'll need a few   extras. &  Restaurants & 3.0 \\ \midrule
Delicious! A perfect brunch spot for   lunch, brunch or dinner. Try the shrimp and grits. &  Restaurants & 5.0 \\
\bottomrule
\end{tabular}
}
\caption{Randomly sampled synthetic reviews generated by the \emph{GPT2} model trained with DP.}
\label{apdx:tbl_gpt2_generations}
\end{table*}

\begin{table*}[!t]
\small
\resizebox{\linewidth}{!}{%
\begin{tabular}{@{}p{0.75\linewidth} p{0.1\linewidth} p{0.1\linewidth} @{}}
\toprule
\textbf{Generated Reviews} & \textbf{\begin{tabular}[c]{@{}l@{}}Business \\ Category\end{tabular}} & \textbf{\begin{tabular}[c]{@{}l@{}}Review\\ Stars\end{tabular}} \\ \midrule
I've   tried a few burgers and it's ok. I don't eat fries (I never do) so don: put   them on your salad or whatever else you have on hand. I have been here many   times for brunch and dinner. & Restaurants & 3.0 \\ \midrule
This   place is one of the best BBQ spots around! They also have many amazing   burgers on the menu. The food is always hot and always tasty. & Bars & 5.0 \\ \midrule
One of   the best concert venues in Reno. Great space and the sound and lighting is   amazing. The sound guys in the stadium really help to get you into the   atmosphere with your music and sound. & Arts \&   Entertainment & 5.0 \\ \midrule
We love   this place. It has a variety of options in the menu, but I always get the   fried chicken which is definitely a better option. If you don't like fried   food, there is a decent selection of regular chicken. You could also choose   to get their bbq, which I am not a fan of, and get a burger. & Restaurants & 3.0 \\ \midrule
Love the   new decor. The new tables are all wood. You don't feel like sitting on an old   bed anymore. They even put their old fireplace on the inside. Food was OK - I   like the steak house. I liked that you can customize the menu to your taste.   The drinks are better too - especially the gin martinis. & Restaurants & 4.0 \\ \midrule
Ordered   a bunch of items, then received confirmation from my Santa that she had   already shipped the items. She did that as I was in the middle of a   drive-thru. When I got home I immediately called the store and asked what the   order was for. They said that they had ordered a lot of stuff (which is nice)   and they wanted to be sure. I said, "Well, what's in it for me?"   They told me it would take a little bit to get out, but when I left they said   they would send me another box. & Shopping & 4.0 \\ \midrule
This   place is a perfect addition to the community. You get a chance to enjoy some   outdoor fun and enjoy all the outdoor activities that you'll find in the   surrounding area. The staff is attentive and professional. It's a great place   to hang out while having a blast. & Arts \&   Entertainment & 4.0 \\ \midrule
I ate   here today. My wife and I were in the area. I ordered the "Gumbo   Sushi". This was a good value considering the size of the bowl. It was   cooked perfectly and the rice was fresh. This place is very well run,   friendly and has a great variety of sushi! & Restaurants & 5.0 \\ \midrule
We went   here to be checked out. I had gone in about 1 1/2 months before. We asked   about getting an appointment and were told they had no one there that could   help us and we just had to go to the front desk and ask. They took care of us   right away. Their nurse was super nice and helped us with our appointment.   She even made sure that we made it into the room without us knowing, and the   COG were there to keep me calm during my appointment which was awesome! I   would highly recommend this place. The room is & Health \&   Medical & 5.0 \\ \midrule
The food   was awesome and friendly. Our server was excellent. I loved that the server   wasn't intrusive with my order. The restaurant was clean and a lot of fun. If   I could make it back here, I would. We will be back next time I'm in Tucson & Restaurants & 5.0 \\ \midrule
I'm not   a fan of Italian cuisine but this was very good. We had the spaghetti and   meatballs, but they were also very tasty. Also had a meatball with bacon on   top. The food is very inexpensive and very authentic, and the atmosphere is   fun and intimate. We will definitely be back! & Restaurants & 5.0 \\ \midrule
Was   expecting a classy place for a casual date night and was disappointed. The   drinks are not worth it. And the service was horrible! & Bars & 1.0 \\ \midrule
We had a   really good time with the team. They were friendly and the service was great.   I had the shrimp tacos which were a total keeper. My boyfriend had his   "Tacos" and he said they were delicious. The chips and salsa were   good too. If your looking for some great local eats in Indy, I highly   recommend this place. & Restaurants & 5.0 \\ \midrule
I was   looking for a spot to meet friends and I came across this beautiful place!   Very quaint and intimate and the service was great as well. Our table was   very small but it was fine as the chairs were just the right height to   comfortably recline. I highly recommend this place. Will definitely be back! & Arts \& Entertainment & 5.0 \\ \midrule
I love   the food here. It's a bit pricey. My wife and I had an amazing experience   there. The place is a great size, it was busy, and we ordered take out. There   was also a server who was kind enough to come over, take our order, etc.   After about 5 minutes, the waitress came back and said she would make our   food for us. This is our first time there, so I think we should make sure we   do not order wrong. We asked for the pork and the rice and she said they were   out of rice & Restaurants & 2.0 \\
\bottomrule
\end{tabular}
}
\caption{Randomly sampled synthetic reviews generated by the \emph{GPT2-Medium} model trained with DP.}
\label{apdx:tbl_gpt2_medium_generations}
\end{table*}

\begin{table*}[!t]
\small
\resizebox{\linewidth}{!}{%
\begin{tabular}{@{}p{0.75\linewidth} p{0.1\linewidth} p{0.1\linewidth} @{}}
\toprule
\textbf{Generated Reviews} & \textbf{\begin{tabular}[c]{@{}l@{}}Business \\ Category\end{tabular}} & \textbf{\begin{tabular}[c]{@{}l@{}}Review\\ Stars\end{tabular}} \\ \midrule
Pleasant experience.   Great food and great service. Good music and the live music really helped   bring out the crowd. Nice, clean place to grab a bite. & Bars & 4.0 \\ \midrule
My boyfriend and I both order the chicken   quesadilla, which comes with 3 pieces of chicken, 2 fried tortillas, sour   cream, rice, and a guacamole. It comes out in about 5 minutes, the tacos are   pretty good and the quinoa is a bit sweet for my taste. Our server was pretty   nice, but was not very friendly or helpful. We're all pretty tired by the   time we get to our table so we didn't want to spend the extra money. I don't   know if my boyfriend got a bad batch of food & Restaurants & 2.0 \\ \midrule
The dentist office at DDS was great. They   were very professional and gave a great service. I've had numerous dental   problems over the years, so I was happy to see that the dentists they employ   are so professional. The only reason I gave them three stars is that there is   no phone calling service to call for follow-up, and their website is so poor   that I couldn't call and they'd have the call placed over an hour later. & Health \& Medical & 3.0 \\ \midrule
One of the best sushi places in the city!   I usually get the chicken and fish roll! It is so fresh and has so much   flavor! The service is excellent. They have a nice selection of beer and   drinks. I highly recommend this place to everyone who loves sushi. & Restaurants & 5.0 \\ \midrule
The food is phenomenal. The portions are   generous. And the service is excellent. & Restaurants & 5.0 \\ \midrule
I'm so glad I tried The Little Noodle.   I've had the chicken curry and the pad thai. It's so good. There was a small   part of me that wanted to try the curry but I was too full. & Restaurants & 5.0 \\ \midrule
My first time at this spot. They were   very friendly and accommodating. The place was clean and the service was   excellent. I will be coming back! I had a burger and fries. & Bars & 5.0 \\ \midrule
Food was really good! I wish they had a   more modern menu but the food is so fresh it would take a long time for me to   go back. Great prices too. & Restaurants & 4.0 \\ \midrule
This place should be called Hotdog King   because of the price. The food wasn't the best, the burgers were ok, but the   whole menu was way too much to consume in one meal. My friend went with her   boyfriend and ordered two different burgers. We ordered the cheesesteak   medium rare. We waited another 5 minutes before the waiter came to take our   food. He took our order and then asked if we wanted our drinks and food   brought out. I didn't realize they only have a microwave and microwave oven.   It wasnt even hot & Hotels \& Travel & 1.0 \\ \midrule
This place is an awesome experience! The   owner and manager were so friendly, friendly and knowledgeable. There were   plenty of great options to choose from and I loved every single meal I had! I   will definitely be returning to this wonderful spot. & Event Planning \& Services & 5.0 \\ \midrule
Food and service was great. Food was just   average and very mediocre. The place was pretty empty, so if you go to check   it out be prepared to wait. & Restaurants & 3.0 \\ \midrule
Just ordered the "special"   platter of 6 shrimp, 5 wings, and a small drink. The platters are big enough   to share, which is a nice touch for two people. & Restaurants & 5.0 \\ \midrule
I'm not sure what happened to these   girls, but every time I walk in and ask for a gel manicure I'm treated with   indifference. I have gone in 3 times and never been offered gel or cuticles   or anything of the kind. It's just a horrible experience that can leave you   feeling very unorganized and unappreciated. I had the worst experience with   two different ladies, both of whom are very nice and have done a great job   with my nails. The third time was very disappointing. Both ladies seemed to   be very frustrated & Beauty \& Spas & 1.0 \\ \midrule
If you want a good Cuban, get the ones in   West Chester. It's always the same thing. Great service, delicious food and a   great price. & Restaurants & 4.0 \\ \midrule
I've been there twice and can't say   enough good things about it. The food was absolutely delicious. We ordered   the "Biscuits" and "Mac \& cheese". I am not sure why   the mac and cheese is a biscuit but it was AMAZING! I would recommend coming   here and eating it as your meal. This is the first time I've tried out this   restaurant and it's definitely my new spot to stop in. & Bars & 5.0
\\
\bottomrule
\end{tabular}
}
\caption{Randomly sampled synthetic reviews generated by the \emph{GPT2-Large} model trained with DP.}
\label{apdx:tbl_gpt2_large_generations}
\end{table*}

\end{document}